\documentclass{article}

\usepackage{arxiv}

\usepackage[utf8]{inputenc} 
\usepackage[T1]{fontenc}    
\usepackage{hyperref}       
\usepackage{url}            
\usepackage{booktabs}       
\usepackage{amsfonts}       
\usepackage{nicefrac}       
\usepackage{microtype}      
\usepackage{lipsum}		
\usepackage{graphicx}
\usepackage{natbib}
\usepackage{doi}
\usepackage{amsmath}
\usepackage{tabularx}
\usepackage{silence}
\title{Information Theory in Open-world Machine Learning Foundations, Frameworks, and Future Directions}

\author{\hspace{1mm}Lin Wang\thanks{Use footnote for providing further
		information about author (webpage, alternative
		address)---\emph{not} for acknowledging funding agencies.} \\
	Shenzhen Key Laboratory of Neuropsychiatric Modulation,\\ Shenzhen-Hong Kong Institute  of Brain Science,\\ Shenzhen Institutes of Advanced Technology, Chinese Academy of  Sciences, Shenzhen 518055, China \\
	\texttt{l.wang@siat.ac.cn} \\
}

\hypersetup{
pdftitle={A template for the arxiv style},
pdfsubject={q-bio.NC, q-bio.QM},
pdfauthor={David S.~Hippocampus, Elias D.~Striatum},
pdfkeywords={First keyword, Second keyword, More},
}

\begin{document}
\maketitle
\begin{abstract}
	Open-World Machine Learning (OWML) aims to develop intelligent systems capable of recognizing known categories, rejecting unknown samples, and continually learning from novel information. Despite significant progress in open-set recognition, novelty detection, and continual learning, the field still lacks a unified theoretical foundation that can quantify uncertainty, characterize information transfer, and explain learning adaptability in dynamic, nonstationary environments. This paper presents a comprehensive review of information-theoretic approaches in open-world machine learning, emphasizing how core concepts such as entropy, mutual information, and Kullback–Leibler divergence provide a mathematical language for describing knowledge acquisition, uncertainty suppression, and risk control under open-world conditions. We synthesize recent studies into three major research axes: information-theoretic open set recognition enabling safe rejection of unknowns, information-driven novelty discovery guiding new concept formation, and information-retentive continuous learning ensuring stable long-term adaptation. Furthermore, we discuss theoretical connections between information theory and provable learning frameworks, including PAC-Bayes bounds, open space risk theory, and causal information flow, to establish a pathway toward provable and trustworthy open-world intelligence. Finally, the review identifies key open problems and future research directions, such as the quantification of information risk, the development of dynamic mutual information bounds, multimodal information fusion, and the integration of information theory with causal reasoning and world-model learning. By bridging fragmented efforts across learning theory, uncertainty quantification, and adaptive AI, this survey provides a unified perspective for building the next generation of self-adaptive, information-aware, and theoretically grounded open-world learning systems.
\end{abstract}


\section{Introduction}
Machine learning has achieved remarkable success in closed-world scenarios\citep{jordan2015,lecun2015}, where all categories, data distributions\citep{pan2009}, and environmental conditions are assumed to be predefined and stationary\citep{gama2014}. However, real-world environments are inherently open: novel classes may emerge\citep{geng2020}, data distributions evolve\citep{zhang2022}, and uncertainty persists\citep{ovadia2019}. These challenges have driven the rapid development of Open-World Machine Learning (OWML), which aims to enable models not only to recognize known categories\citep{han2020,han2019} but also to reject and learn from the unknown\citep{parmar2023}. Although OWML integrates areas such as open set recognition\citep{scheirer2014}, novelty discovery\citep{cao2021}, and continual learning\citep{wang2024}, the theoretical understanding of how knowledge and uncertainty evolve in open environments remains limited\citep{zhu2024}.

Most existing approaches to open-world learning rely on heuristic confidence thresholds\citep{chen2021}, empirical energy models\citep{liu2020}, or incremental memory-based learning\citep{rebuffi2017}. These methods perform well in practice, but lack a unified mathematical foundation that can explain why and when a system can safely recognize, reject, and adapt. The absence of such a foundation makes it difficult to quantify the limits of adaptability, control the risks of unknown exposure, or provide theoretical guarantees for the stability of continual learning. Therefore, establishing a generalizable and provable framework for OWML has become a fundamental open problem in the field.

Information theory provides a powerful lens for addressing this challenge. Its core quantities, entropy, mutual information\citep{shannon1948}, and Kullback-Leibler divergence, naturally measure uncertainty\citep{kullback1951}, knowledge transfer, and information compression\citep{varley2023}. By interpreting the learning process as an information flow from input to representation to decision\citep{weimar2025,tan2023}, information theory allows the dynamics of open-world learning to be described and analyzed mathematically consistent\citep{dahlke2025,hu2024}. This perspective unifies multiple tasks, from recognizing the known and rejecting the unknown to continually learning the novel\citep{mondal2025,wen2025}, under a single information-theoretic formulation.

Recent work has begun to explore this direction. In particular,some research introduced an information-theoretic formulation of open-world learning, interpreting the balance between knowledge retention and unknown suppression as an optimization of mutual information\citep{dziugaite2018}. Subsequent studies extended this idea to continuous learning under open-world distribution shifts\citep{tan2024}. These advances suggest that information theory may serve as the missing theoretical backbone for understanding and formalizing open-world intelligence.

In this review, we provide the first systematic synthesis of information-theoretic approaches to open-world machine learning. We revisit classical and modern theories of information, analyze their roles across major OWML tasks, and identify the key open problems that remain. We aim to bridge fragmented studies across open-set recognition, novelty discovery, and continual learning into a unified framework that connects empirical progress with theoretical foundations. Finally, we outline open challenges and emerging trends, including information-risk quantification, dynamic mutual information bounds, and the integration of information theory with causality and world models.
\section{Background: Information Theory and Open-world Machine Learning}
\label{sec:headings}

Machine learning operates fundamentally as a process of information acquisition, compression, and transmission\citep{kawaguchi2023,chen2025}. While traditional learning theories assume closed and stationary environments, real-world intelligence must continually adapt to uncertainty, novelty, and change\citep{kim2025,wang2024}. This section provides the conceptual and theoretical background for understanding open-world machine learning through the lens of information theory. We first review the essential principles of information theory, then summarize the foundations and challenges of open-world learning, and finally discuss the conceptual intersection between the two paradigms.

\subsection{Fundamentals of Information Theory in Learning Systems}
Information theory, founded by Claude Shannon\citep{shannon1948}, provides a rigorous mathematical framework for quantifying uncertainty and information flow. Its three central quantities—entropy, mutual information, and Kullback–Leibler (KL) divergence—form the backbone of how knowledge, uncertainty, and learning dynamics are measured in modern machine learning systems\citep{kullback1951}.

Entropy\citep{bulte2025,wood2024,zhang2025} represents the average level of uncertainty in a random variable, making it an essential tool for measuring unpredictability in data, model predictions, or decision boundaries. High entropy indicates uncertainty, while low entropy reflects confident or deterministic predictions.
Mutual information (MI)\citep{dahlke2025,tsur2023} quantifies the amount of information shared between two variables, capturing the dependency between inputs, representations, and outputs. In representation learning, mutual information measures how much of the input data’s relevant structure is preserved in the learned representation.
KL divergence\citep{flynn2023,kuzborskij2024,roulet2025}, a measure of discrepancy between probability distributions, expresses how much one belief or model deviates from another—such as how much a posterior differs from a prior after learning.

These measures collectively describe the information flow that underlies learning. A learning system can be viewed as a communication channel X→Z→Y, where the input data X is encoded into a latent representation Z that supports the prediction of task labels Y. The goal of learning is to maximize task-relevant information I(Z;Y) while minimizing redundant or noisy information I(Z;X). This trade-off forms the basis of the Information Bottleneck (IB) principle proposed by\citep{tishby2000}.
The IB principle reframes learning as a compression–relevance optimization problem: the model should compress the input while preserving information useful for predicting the output. Subsequent developments, such as the Deep Variational Information Bottleneck\citep{alemi2016} and Information-theoretic Generalization Bounds\citep{negrea2019,neu2021}, have further demonstrated that information quantities can explain generalization, robustness, and adaptation in modern deep learning systems. These foundational ideas provide the theoretical grounding for extending information-theoretic reasoning to open-world environments.
\subsection{The Foundations of Open-world Machine Learning}
Traditional machine learning operates under the closed-world assumption, where all possible classes and data distributions are known and fixed during training and inference. In contrast,OWML relaxes this assumption, confronting models with the challenges of novelty, uncertainty, and non-stationarity\citep{rios2024,liu2023ai,mundt2023wholistic}.

OWML envisions intelligent systems capable of recognizing known categories\citep{federici2020}, rejecting unfamiliar inputs\citep{ghassemi2022,vaze2022}, and continually acquiring new knowledge\citep{de2021continual}. This paradigm integrates three closely related research areas. The first is open-set recognition\citep{ge2017generative,wang2023glocal}, which focuses on identifying and safely rejecting samples that do not belong to any known category. The second is novelty discovery\citep{Han_2022,zhang2022automatically}, which involves grouping and interpreting previously unseen samples into coherent new classes. The third is continual learning\citep{de2021continual,mirzadeh2020understanding}, which enables models to assimilate new information while retaining previously acquired knowledge, thereby avoiding catastrophic forgetting. Collectively, these three components constitute the fundamental cycle of open-world learning, in which a model must first reject unknown data, then discover new concepts, subsequently learn them, and finally integrate them into its existing knowledge base.

However, OWML introduces several fundamental theoretical challenges. First, the label space is no longer fixed\citep{chen2018lifelong}, meaning the hypothesis space evolves as the environment changes. Second, data distributions become non-stationary\citep{farquhar2019unifying,zhou2022domain,liang2025comprehensive,kurle2019continual}, making classical risk minimization assumptions invalid. Third, uncertainty becomes multi-faceted—originating from unknown classes\citep{boult2019learning}, shifting distributions\citep{thiagarajan2022}, and partial observability\citep{jafarzadeh2022}.

Existing approaches often rely on heuristic solutions such as confidence thresholds\citep{wang2021energy,liu2020simple}, distance-based novelty scoring\citep{cheng2021learning}, or memory replay strategies\citep{wang2022continual,raghavan2019}. While effective empirically, these methods lack formal guarantees regarding stability\citep{sun2025textttrtext}, adaptability\citep{bonjour2024towards,tang2025similarity}, or safety\citep{mohseni2022taxonomy}. The absence of a unified theoretical framework—one that can quantify how information about the world evolves—remains the central obstacle to developing robust open-world intelligence\citep{xue2024advancing}.
\subsection{Bridging Information Theory and Open-world Learning}
Information theory offers a natural mathematical foundation for formalizing open-world learning\citep{tishby2015deep,kejriwal2024challenges}. It provides a set of distribution-agnostic tools\citep{kejriwal2024challenges,chen2021generalization} to quantify uncertainty\citep{fakour2024structured}, learning progress\citep{li2024robust}, and adaptation cost\citep{nguyen2021information}, thus addressing many of the limitations of existing OWML frameworks\citep{xu2023ibrarin,sun2021m2iosr,wutschitz2023rethinking}. From an information-theoretic viewpoint, learning in open environments can be conceptualized as a process of information flow under uncertainty\citep{rios2024uncertainty,gawlikowski2023survey}, where models must decide how much information to extract, retain, and suppress\citep{achille2018information}.

Entropy provides a direct measure of uncertainty in the presence of unknown categories\citep{wang2023entropy}. Mutual information quantifies how effectively a model preserves useful knowledge about known tasks\citep{westphal2024information}, while KL divergence captures the adaptation cost when encountering novel distributions\citep{nguyen2021kl}. This framework allows the behaviors of recognition, rejection, and adaptation to be expressed in unified informational terms\citep{sun2023graph}.

Recent studies have begun to explore this intersection explicitly. \citep{li2025exploring}formulated open-world learning as an information-optimization problem, balancing knowledge retention and unknown suppression through mutual information.\citep{li2025improving}extended this to continual learning, modeling open-world adaptation as a dynamic flow of information across time. These approaches demonstrate that information theory is not merely a descriptive tool but a foundational theory for analyzing, quantifying, and even proving properties of open-world learning systems.

Ultimately, the convergence of information theory and OWML reframes the open-world learning challenge from an empirical problem into a quantifiable and provable information process. It enables researchers to reason about learning not only in terms of accuracy or error but also in terms of information gain, uncertainty reduction, and knowledge evolution. This synthesis lays the groundwork for the next section, which introduces formal information-theoretic frameworks for open-world machine learning.
\section{Information-Theoretic Framework for Open-world Machine Learning}
OWML requires learning systems to operate under uncertainty, evolving environments, and continually expanding label spaces\citep{zhang2025ai,du2023bridging}. Traditional empirical risk minimization frameworks assume a fixed distribution and a closed hypothesis space, making them insufficient for reasoning about the dynamics of open environments\citep{boult2019learning,xie2024deep}. Information theory offers an elegant and mathematically grounded alternative: it interprets learning as an information flow process\citep{lakkaraju2017identifying}, where knowledge is acquired, compressed, and transmitted through representations that balance relevance, uncertainty, and adaptability\citep{cruz2025open}. This section introduces the information-theoretic framework that formalizes OWML as a quantifiable process of information exchange.
\subsection{The Information Flow Perspective of Open-world Learning}
In open-world environments, a learning system continuously interacts with uncertain information sources, transforming raw sensory data into structured knowledge. From an information-theoretic perspective, this process can be represented as an information channel connecting three conceptual spaces: the input space X, the representation space Z, and the output space Y. The input data contain both task-relevant and irrelevant information, and the learning objective is to filter and compress this information such that Z preserves only what is useful for predicting Y.

This perspective directly extends the Information Bottleneck (IB) principle proposed by\citep{tishby2000}, which formulates the learning problem as an optimization of mutual information:
\begin{equation}
\min I(Z; X) - \beta I(Z; Y),
\label{eq:ib}
\end{equation}
where $I(\cdot;\cdot)$ denotes mutual information and $\beta$ controls the balance between input compression and task relevance.

The IB framework interprets learning as an optimal encoding problem, in which a compact representation Z should retain only the information about X that is predictive of Y. This view provides a natural foundation for modeling the learning process as an information flow, bridging the gap between representation learning and decision-making.

In the open-world context, however, this classical formulation must be extended to account for information associated with unknown or novel categories\citep{bendale2016towards}. Unlike closed-world settings, where all possible outputs are predefined, open-world systems must regulate the information they extract from uncertain or previously unseen inputs\citep{dhamija2018reducing,zhao2023revisiting}. This requirement motivates the development of an extended information-theoretic objective for OWML.
\subsection{The Information-Theoretic Objective for OWML}
\citep{zhou2024continuous} proposed an information-theoretic formulation that generalizes the Information Bottleneck to open-world learning scenarios.
The central insight is that open-world learning involves managing three types of information:
(1) compressing redundant information from the input,
(2) retaining task-relevant information for known categories, and
(3) suppressing misleading information from unknown or uncertain samples.
This trade-off is expressed as an optimization problem over mutual information quantities:
\begin{equation}
\min I(Z; X) - \beta I(Z; Y_{\text{known}}) + \gamma I(Z; Y_{\text{unknown}}),
\label{eq:owml}
\end{equation}
where $I(Z; X)$ represents the compression of the input data into a latent representation $Z$;
$I(Z; Y_{\text{known}})$ quantifies the task-relevant information preserved for known classes;
and $I(Z; Y_{\text{unknown}})$ measures the influence of unknown data on the learned representation.
The parameters $\beta$ and $\gamma$ control the trade-off between knowledge retention and novelty suppression.

This formulation unifies the three fundamental aspects of open-world learning—compression, retention, and rejection—into a single mathematical objective.
It extends the classical IB framework by incorporating a mechanism for uncertainty regulation: rather than assuming a fixed output distribution, it explicitly models the influence of unknown categories as an information source to be minimized.
This objective thus provides a foundation for understanding safe adaptation, where a model can learn effectively from known data while avoiding overconfidence on unknown inputs.

Building upon this formulation, \citep{li2025improving} extended the objective to continual learning under open-world distribution shifts.
They proposed that knowledge retention over time can be represented as the preservation of mutual information between successive representation states:
\begin{equation}
\max_t I(Z_t; Z_{t-1}),
\label{eq:continual}
\end{equation}
where $Z_t$ denotes the latent representation at time step $t$.
Maintaining high mutual information between $Z_t$ and $Z_{t-1}$ ensures that new learning does not catastrophically overwrite previously acquired knowledge.

This dynamic regularization term introduces a temporal dimension to the information-theoretic framework, allowing OWML to capture both stability and adaptability across evolving environments.

Together, these formulations transform open-world learning from an empirical problem into a quantifiable process of information regulation.
Learning becomes not merely a matter of minimizing error but of managing the balance between information compression (generalization), information preservation (memory), and information suppression (safety).

\subsection{Extensions and Theoretical Connections}
The information-theoretic formulation of OWML is closely connected to several established theoretical frameworks, each of which provides complementary insights.

Relation to the Information Bottleneck (IB):
The OWML objective generalizes the IB principle by adding a term for novelty suppression\citep{xu2019open}. While the IB focuses solely on balancing compression and relevance, the OWML formulation introduces an additional constraint to control uncertainty caused by unknown data. This extension allows the model to operate safely beyond the closed-world assumption.

Relation to PAC-Bayes and Generalization Theory:
Mutual information plays a central role in bounding generalization error. \citep{haghifam2020sharpened} showed that the expected generalization gap can be upper-bounded by a function of the mutual information between training data and model parameters. Within OWML, limiting the information a model extracts from unknown inputs ensures tighter, provable bounds on generalization under distribution shifts.

Relation to Open-space Risk Theory:
\citep{scheirer2012toward} introduced open-space risk to measure the danger of making confident predictions in unsupported regions of feature space. In information-theoretic terms, these regions correspond to high-entropy, low-mutual-information areas where models should minimize exposure. The OWML framework thus reinterprets open-space risk as an information exposure problem.

Relation to Continual and Lifelong Learning:
In continual learning, the mutual information preservation objective $I(Z_t; Z_{t-1})$ serves as a theoretical expression of stability–plasticity balance\citep{chen2023stability}. Retaining mutual information across tasks helps maintain previously learned representations, while the addition of controlled new information enables adaptive knowledge expansion.

These theoretical connections collectively position information theory as the backbone of open-world learning. It provides a unifying language for describing diverse learning phenomena—recognition, rejection, and adaptation—in terms of quantifiable information dynamics.

\section{Applications of Information Theory in OWML Subtasks}

Information theory not only provides a unifying mathematical foundation for understanding open-world learning but also offers practical tools for analyzing and improving its key subtasks\citep{liu2025learning}. This chapter examines how entropy, mutual information, and divergence-based measures have been applied in three major areas of OWML: open-set recognition, novelty discovery, and continual learning.
Each of these subtasks addresses a different phase of the open-world cycle—recognizing the known, discovering the novel, and integrating the new—yet all can be expressed within a shared information-theoretic framework.

\subsection{Information Theory in Open-set Recognition (OSR)}

Open-set recognition (OSR) is the first and most fundamental task in open-world learning\citep{cao2025survey,nawaz2025beyond}. Its goal is to enable models to correctly classify known classes while rejecting samples that belong to unseen or unknown categories\citep{xing2025towards,moazzami2025open}. The key challenge is balancing discriminative capability with uncertainty awareness—how to separate known information from informational noise introduced by the unknown.

From an information-theoretic standpoint, OSR can be viewed as minimizing the entropy of known predictions while maximizing the uncertainty for unknown samples\citep{liu2025information}.
Entropy serves as a natural metric of confidence: low entropy indicates certainty about known classes, whereas high entropy signals uncertainty that can be used to reject unfamiliar inputs\citep{garg2022domain}.
Formally, OSR can be formulated as an entropy-regularized optimization problem\citep{hu2024information}:
\begin{equation}
\min \mathbb{E}_{x \in \mathcal{D}_{\text{known}}} [H(Y|X)] 
\; + \; \lambda \, \mathbb{E}_{x \in \mathcal{D}_{\text{unknown}}} [ - H(Y|X) ],
\label{eq:osr}
\end{equation}
where $H(Y|X)$ is the conditional entropy of the output given the input,
and $\lambda$ balances classification confidence and rejection safety.
Beyond entropy, mutual information has been used to improve rejection decisions.
By maximizing $I(Z; Y_{\text{known}})$ while minimizing $I(Z; Y_{\text{unknown}})$, the model learns feature representations that retain discriminative information for known classes but discard information that correlates with unknowns.
This information separation principle underpins recent OSR models such as OpenMax\citep{bendale2016towards}, ARPL\citep{chen2021}, and M2IOSR\citep{sun2021m2iosr}).
Information-theoretic OSR therefore provides both an intuitive and quantitative measure of model confidence and generalization beyond the closed set.

\subsection{Information Theory in Novelty Discovery}

Once unknown samples are identified, the next step in open-world learning is novelty discovery—the process of grouping, characterizing, and interpreting unknown data to form new classes or concepts\citep{jin2024knownnovel}. This task lies at the intersection of unsupervised learning and knowledge expansion\citep{zhou2022open}, where the model must decide what constitutes “novel” information.

In information-theoretic terms, novelty discovery can be formalized as a maximization of information gain, i.e., the difference between the entropy before and after observing new data\citep{du2023can,lidayan2025intrinsically}. The objective is to identify clusters or representations that maximize the reduction in uncertainty about the environment\citep{jafarzadeh2020review}:
\begin{equation}
\max_{\theta} \; \mathbb{E}_{x \in \mathcal{D}_{\text{unknown}}} 
\left[ H(P_{\text{prior}}(Y)) - H(P_{\theta}(Y|X)) \right],
\label{eq:novelty}
\end{equation}
where $H(P_{\text{prior}}(Y))$ denotes prior uncertainty and $H(P_{\theta}(Y|X))$ denotes posterior uncertainty after modeling with parameters $\theta$.
The term inside the brackets represents information gain—a measure of how much new information about the world has been discovered.

This information gain formulation aligns with the principles of active learning and Bayesian exploration\citep{houlsby2011bayesian}, where models seek data that maximally reduce uncertainty\citep{sekar2020planning}.
In practice, novelty discovery methods often combine mutual information with clustering algorithms (e.g., InfoNCE\citep{oord2019representation}, Deep InfoMax\citep{hjelm2019learning}) to identify meaningful latent structures in the unknown data space.
Recent research \citep{abbasi2024deciphering,wang2024hilo} extends this idea by coupling mutual information maximization with representation disentanglement, ensuring that novel knowledge is structured and separable from known features.

\subsection{Information Theory in Continual Learning}

Continual learning (CL) addresses the problem of learning from a sequence of tasks without catastrophic forgetting\citep{wang2024comprehensive}. In open-world environments, this challenge becomes even more critical as new tasks and distributions emerge continuously\citep{li2025exploring}. Information theory provides a principled approach to analyze and mitigate forgetting by quantifying how much information about past knowledge is retained as new learning occurs\citep{song2023infocl}.

\citep{li2025estimating} formalized this idea by defining mutual information preservation across time. Their objective encourages the model to maintain shared information between consecutive latent representations, ensuring temporal stability in formulation~(\ref{eq:continual}), this formulation provides a compact theoretical explanation for methods such as EWC\citep{kirkpatrick2017}, LwF\citep{li2017learning}, and replay-based continual learning\citep{wen2025information}, which implicitly preserve information through regularization or memory mechanisms.

Additionally, the trade-off between plasticity (learning new information) and stability (retaining old knowledge) can be expressed as a dual optimization problem over mutual information:
\begin{equation}
\max \; I(Z_t; Y_t) \; - \; \lambda \, D_{\text{KL}}(P(Z_t) \parallel P(Z_{t-1})),
\label{eq:stability}
\end{equation}
where the first term promotes adaptation to the current task,
and the KL-divergence term penalizes excessive deviation from previous representations.
This equation directly connects continual learning to the information-theoretic framework of OWML, showing that maintaining controlled information divergence is essential for sustainable knowledge evolution.

Through this perspective, continual learning becomes an information balancing process—preserving relevant past information while incorporating new, task-specific knowledge.

\subsection{Summary of Information-Theoretic Applications}
The integration of information-theoretic principles into open-world learning tasks provides a consistent mathematical foundation for reasoning about uncertainty, discovery, and retention.
Information theory thus acts as a bridge across all components of OWML:
entropy quantifies uncertainty, mutual information measures knowledge retention, and divergence captures adaptation cost.
Together, these measures transform open-world learning into a unified, quantifiable process of information flow and evolution.

\section{Toward Provable Open-world Learning:Mathematical Foundations}

OWML aims to build systems that can operate safely and adaptively under uncertainty. To transition from empirical success to provable intelligence, OWML requires a rigorous mathematical foundation. This chapter explores how information theory, combined with generalization theory, causal inference, and statistical learning principles, can be used to construct provable learning guarantees under open-world conditions. We begin by revisiting classical closed-world theories, extend them to open distributions, formalize information-theoretic risk and bounds, and discuss future directions toward provable open-world intelligence.

\subsection{Revisiting Closed-world Provability}

Traditional machine learning operates under a closed-world assumption, where data are drawn i.i.d. from a fixed distribution $P(X, Y)$, and all classes are known during both training and inference\citep{zhang2025ai}. In this setting, the generalization error—the difference between empirical risk on the training set and true risk on unseen samples—can be bounded using statistical learning theory\citep{yu2021understanding}.

A classical guarantee\citep{gross2025improved} can be expressed as:
\begin{equation}
\mathcal{R}(f) \leq \hat{\mathcal{R}}(f) + \mathcal{C}(f, n, \delta),
\label{eq:generalization}
\end{equation}
where $\mathcal{R}(f)$ denotes the expected risk, $\hat{\mathcal{R}}(f)$ is the empirical risk, and $\mathcal{C}(f, n, \delta)$ is a complexity term depending on model capacity, sample size $n$, and confidence parameter $\delta$.

While this formulation provides solid theoretical grounding for closed-world learning, it collapses under distributional openness—when new classes or data distributions emerge\citep{chuang2020estimating}. The core limitation is that traditional generalization bounds assume a fixed hypothesis space and a stationary data-generating process, both of which are violated in open-world settings\citep{feldman2019}.

\subsection{Extending Generalization to Open Distributions}

In OWML, the learner must adapt to new data domains and evolving label spaces, 
where the training distribution $P_{\text{train}}(X, Y)$ differs from the test distribution $P_{\text{test}}(X, Y)$\citep{stojanov2021domain}. The challenge is thus to establish \textit{transferable} or \textit{adaptive} generalization guarantees.

A common approach is to introduce distributional divergence measures that quantify the difference between source and target distributions\citep{courty2016optimal}. 
Let $D_{\mathrm{KL}}(P_{\text{test}} \parallel P_{\text{train}})$ denote the Kullback--Leibler divergence. The expected risk under open-world conditions can be upper-bounded as\citep{JMLR:v20:17-750}:
\begin{equation}
\mathcal{R}_{\text{test}}(f) 
\leq \mathcal{R}_{\text{train}}(f)
+ \alpha \, D_{\text{KL}}(P_{\text{test}} \parallel P_{\text{train}}),
\label{eq:open_bound}
\end{equation}
where $\alpha$ is a scaling constant that captures the model’s sensitivity to distributional shifts. This inequality reflects a fundamental insight: the cost of openness can be quantified as the information divergence between what has been learned and what is newly encountered.

In practice, this divergence term can be estimated or regularized through mutual information constraints, effectively linking distributional generalization with information-theoretic control.

\subsection{Information-theoretic Risk and Provable Bounds}

Information theory provides a natural way to express provable guarantees through mutual information-based risk bounds\citep{xu2017information}.
Building on the PAC-Bayes framework, \citep{harutyunyan2021} demonstrated that the generalization error can be bounded in terms of the mutual information between the model parameters $W$ and the training data $D$:
\begin{equation}
\mathbb{E}\left[\mathcal{R}(f_W) - \hat{\mathcal{R}}(f_W)\right]
\leq \sqrt{\frac{2 I(W; D)}{n}}.
\label{eq:pacbayes}
\end{equation}
This inequality provides a powerful interpretation: models that encode less information about the specific training data generalize better.
In open-world learning, this concept extends to controlling the information exchanged between past knowledge, current tasks, and novel environments.

Combining this with the open-world formulation from Chapter 3, we can express the Information-Theoretic Open-world Bound as:
\begin{equation}
\mathcal{R}_{\text{open}}(f)
\leq \hat{\mathcal{R}}_{\text{known}}(f)
+ \sqrt{\frac{2 I(Z; X_{\text{known}})}{n}}
+ \gamma D_{\text{KL}}(P_{\text{unknown}} \parallel P_{\text{known}}),
\label{eq:ow_bound}
\end{equation}
where the first term is the empirical risk on known data, the second term quantifies uncertainty and generalization ability through mutual information, and the third term penalizes deviation caused by unknown distributions.
This provides a mathematically grounded definition of open-world risk—the total uncertainty of learning under novelty and distributional drift.

\subsection{Toward Provable Open-world Intelligence}

Developing provable open-world learning requires more than empirical adaptation; it demands formal principles that link uncertainty, adaptability, and stability\citep{qu2025dual}. The integration of information theory into statistical learning opens several promising research avenues:

Information Risk Theory:
Establishing information risk minimization (IRM) as a generalization of empirical risk minimization (ERM).
In IRM, the objective is to minimize the expected information exposure to unknown factors, rather than just empirical error.

Dynamic PAC-Bayes Bounds:
Extending PAC-Bayes inequalities to non-stationary and temporally evolving tasks, where the mutual information term becomes time-dependent $I(W_t; D_{t-1})$.
This could provide guarantees for continual learning under open-world dynamics.

Causal Information Flow:
Incorporating causal reasoning into information-theoretic analysis to distinguish between informative novelty (causally relevant) and spurious novelty (noise).
This would enable provable causal generalization in open-world systems.

Information Stability and Safety:
Defining safety margins in terms of information bounds—for example, ensuring that  $D_{\mathrm{KL}}(P_{\text{unknown}} \parallel P_{\text{known}}) < \epsilon$ for safe model deployment.
Such criteria could be used to certify when an open-world model is theoretically safe to operate.

Bridging Theory and Implementation:
Translating these bounds into practical algorithms via information regularization, adaptive loss functions, and mutual information estimators.
This would allow theory-driven model design where stability, adaptability, and uncertainty are mathematically coupled.

Collectively, these directions move open-world learning toward provable intelligence—systems that not only perform well empirically but also offer verifiable guarantees about their behavior under novel and uncertain conditions.
Such a theory would mark the transition from heuristic adaptation to quantified, principled open-world reasoning.

\section{Comparative Theoretical Analysis}

Classical theories such as Statistical Learning Theory\citep{bartlett2020}, Bayesian Learning\citep{wilson2020bayesian}, Energy-based Modeling\citep{song2021train}, and Causal Learning\citep{kaddour2022causal} each capture different aspects of intelligence—generalization, uncertainty, representation, and interpretability.
However, they are all built upon assumptions of closure, stability, or complete knowledge.This chapter presents a comparative theoretical analysis showing how information theory provides a unified foundation that extends, connects, and generalizes these paradigms under open-world conditions.

\subsection{Statistical Learning Theory}

Statistical Learning Theory (SLT) provides the mathematical foundation for generalization in closed-world settings. Its strength lies in its provable nature: it defines clear relationships between empirical performance\citep{boucheron2005}, model capacity, and expected risk\citep{feldman2019}. This framework assumes that all samples are independent and identically distributed and that the underlying data distribution remains fixed.
While SLT has been extraordinarily influential in defining what it means for a model to generalize, it collapses under openness. When new classes appear\citep{geng2020} or data distributions shift\citep{machlanski2025}, the assumption of a stationary environment no longer holds, and classical risk bounds become invalid\citep{wiles2021fine}.
From an information-theoretic perspective, SLT can be informationally extended.
Information theory transforms risk into an information exchange process, where the uncertainty between model representations and data distributions is explicitly quantified. This shift allows learning guarantees to remain meaningful even when the world is not fixed, providing a bridge between closed-world provability and open-world adaptability.
In summary, Statistical Learning Theory remains the formal backbone of provability, but its extension through information measures such as entropy and mutual information enables it to survive in dynamic, open settings.

\subsection{Bayesian Learning}

Bayesian Learning introduces a probabilistic framework for managing uncertainty\citep{wilson2020bayesian}. Its advantage lies in its strong ability to represent epistemic uncertainty—the uncertainty about model parameters or hypotheses. By maintaining a distribution over model parameters, Bayesian approaches allow reasoning under incomplete information and offer principled probabilistic inference.
However, Bayesian learning relies on predefined priors and fixed model structures\citep{li2021detecting}. In open-world conditions, priors may become outdated, incomplete, or even misleading as the environment evolves\citep{adel2025bayesian}. This makes traditional Bayesian inference brittle when confronted with novel categories or shifting semantics\citep{galashov2024non}. Information theory complements Bayesian learning by embedding uncertainty management directly into information flow. Instead of assuming fixed priors, information-theoretic learning models uncertainty as a dynamic process of encoding, transmission, and transformation of information. Mutual information becomes the connecting principle—linking the belief-based uncertainty of Bayesian inference to the relevance-based uncertainty of open-world adaptation.

In summary, Bayesian learning interprets uncertainty as belief; information theory interprets it as information relevance. Together, they form a foundation for reasoning under both known and unknown uncertainties.

\subsection{Energy-based Models}

Energy-based Models (EBMs) have achieved remarkable empirical success, particularly in deep learning\citep{oliva2025uniform}. They define a scalar “energy” function that measures how compatible a configuration of variables is, allowing for flexible modeling of complex dependencies\citep{xu2024energy}. EBMs are powerful because they can capture high-dimensional relationships without requiring explicit probability normalization, making them practical and expressive\citep{song2021train}.
However, EBMs are largely empirical and lack a comprehensive theoretical framework explaining their learning dynamics, stability, or generalization. Their energy function can be interpreted as an implicit potential landscape, but its theoretical meaning often remains opaque. From the viewpoint of information theory, energy can be reframed as information potential—a representation of how much information is stored or compressed in a given state.
Under this interpretation, minimizing energy becomes equivalent to optimizing the information flow within the system, aligning EBMs with information-theoretic principles of efficiency and stability.
This provides a theoretical bridge: information theory endows energy-based learning with measurable interpretability, connecting it to the entropy and mutual information structures underlying open-world intelligence.

In summary, EBMs offer practical performance, while information theory offers them a missing theoretical backbone—turning energy minimization into an explicit form of information optimization.

\subsection{Causal Learning}

Causal Learning seeks to uncover and utilize the cause–effect relationships underlying observed data. Its advantage lies in its interpretability and its ability to generalize across interventions and domains\citep{jiao2024causal}. Causal inference assumes that the structural relationships between variables remain invariant, allowing predictions to hold even under changes in external conditions.
However, causal learning depends on having access to stable causal structures.
When new causes emerge or mechanisms evolve—as in open-world environments—this assumption no longer holds. Traditional causal frameworks struggle to adapt to causal novelty or explain newly arising dependencies.
Information theory extends causal reasoning by introducing the concept of information–causal flow. This framework treats causation as an information transmission process, where the strength and direction of influence are quantified through changes in information content.
By viewing causal links as dynamic information channels, information theory enables models to detect when existing mechanisms break and new ones form.
This creates a pathway toward adaptive causal inference, where causality and novelty are jointly represented in the information space.
In summary, causal learning ensures interpretability under stable mechanisms;
information theory generalizes it to handle causal openness, forming the basis for adaptive, self-updating causal systems.

\subsection{Comparative Theoretical Summary}

The comparison across the four paradigms reveals that information theory acts as a meta-theoretical bridge—linking provability, uncertainty, representation, and causality under a unified formalism of information flow.
Where classical theories assume closure and stability, information theory reinterprets them as special cases of information transformation within evolving environments.
\begin{table}[ht!]
\centering
\caption{Comparative analysis of major theoretical frameworks and their relationship with Information Theory in OWML}
\begin{tabularx}{\textwidth}{lXXX}
\toprule
\textbf{Theoretical Framework} & \textbf{Advantages} & \textbf{Limitations} & \textbf{Relation to Information Theory} \\
\midrule
\textbf{Statistical Learning Theory} & Theoretically complete and provable & Relies on closed-world distribution assumptions, unable to handle open-world dynamics & Can be extended through information measures to provide provable bounds under open distributions \\
\textbf{Bayesian Learning} & Strong ability to handle uncertainty & Sensitive to prior assumptions and vulnerable to novel classes & Can be integrated with mutual information to model dynamic uncertainty \\
\textbf{Energy-based Models} & Excellent empirical performance & Lack rigorous theoretical interpretation & Can be viewed as Information Potential Models, where energy reflects informational capacity \\
\textbf{Causal Learning} & Strong interpretability and transferability & Depends on stable structures, difficult to adapt to new causal mechanisms & Can be extended to Info-Causal Flow, integrating causality with information transmission \\
\bottomrule
\end{tabularx}
\end{table}
Statistical learning offers provability, Bayesian learning offers uncertainty reasoning, energy-based models offer practical expressiveness, and causal learning offers interpretability.
Yet all rely on the assumption of a closed and stationary world.
Information theory unifies these paradigms under a single conceptual lens: learning as a process of dynamic information regulation.
It redefines risk as information transfer, uncertainty as information entropy, energy as information potential, and causality as information flow.
Through this perspective, OWML transforms machine learning from a static optimization paradigm into a living system of information exchange, capable of reasoning, adapting, and proving its behavior in open and uncertain worlds.

\section{Open Problems and Future Directions}

OWML remains far from a mature theoretical discipline. Information theory has provided a powerful foundation for reasoning about uncertainty, adaptability, and provability, but many scientific questions remain unanswered. This chapter outlines key open problems and proposes several promising research directions that could define the next stage of development for information-theoretic open-world learning.
Each direction highlights a core challenge, its underlying scientific question, and a potential conceptual goal for future exploration.

\subsection{Open Problems Overview}

The following table summarizes the principal future directions and open theoretical challenges in information-theoretic OWML.
\begin{table}[ht!]
\centering
\caption{Future research directions and open problems in information-theoretic Open-world Machine Learning}
\begin{tabularx}{\textwidth}{lXX}
\toprule
\textbf{Direction} & \textbf{Scientific Question} & \textbf{Potential Research Objective} \\
\midrule
Information Risk Theory & How to define open information risk under non-stationary and uncertain conditions & Establish an Open-space Information Risk framework for quantifying information exposure in open environments \\
Dynamic Information Flow & How do information generalization bounds evolve over time in continual and incremental learning scenarios & Develop Temporal Mutual Information formulations for time-dependent information retention and adaptation \\
Information and Causality & How can information flow explain or predict causal flow in evolving systems & Build an Info-Causal Fusion model that unifies information theory with dynamic causal inference \\
Multimodal Information Theory & How can mutual information across different modalities be measured and combined effectively & Define Cross-modal Information Bounds for aligning heterogeneous sensory and symbolic representations \\
Self-adaptive Agents & Can an intelligent agent quantify and regulate its own information boundary & Design Self-Information-Aware Agents capable of evaluating and optimizing their internal informational states \\
\bottomrule
\end{tabularx}
\end{table}

\subsection{Directional Analysis}

\subsubsection{Information Risk Theory}
One of the most fundamental open problems is how to define and measure information risk in open-world learning. Traditional risk is expressed in terms of error or loss under fixed distributions, but in open environments, models face not only predictive errors but also information exposure—how much of the unknown world they fail to represent or misinterpret. A future Open-space Information Risk framework would quantify this exposure as a balance between known, unknown, and novel informational components, providing a new axis of theoretical provability for open systems.
\subsubsection{Dynamic Information Flow}
Most existing information-theoretic bounds, such as mutual information constraints, are static and assume a single learning stage. In open-world scenarios, however, learning unfolds over time: information is accumulated, transformed, and sometimes forgotten. Understanding how generalization and retention evolve dynamically requires formulating Temporal Mutual Information—a time-dependent measure capturing how information propagates and stabilizes across sequential tasks. This direction could lay the groundwork for a dynamic, provable continual learning theory.
\subsubsection{Information and Causality}
A long-term scientific challenge lies in merging causal reasoning with information-theoretic flow. While causality explains why events occur, information theory explains how signals about those events are transmitted and processed.
The question of whether information flow can serve as a sufficient descriptor for causal flow remains open. A unified Info-Causal Fusion framework could describe both mechanisms and communication processes in adaptive systems, allowing models to infer not only correlations but evolving causal structures in open environments.
\subsubsection{Multimodal Information Theory}
As real-world systems increasingly rely on multimodal data—text, vision, speech, sensors—the challenge is to define mutual information consistently across modalities that differ in structure and scale. Information theory provides a natural bridge, yet cross-modal dependencies are often nontrivial and asymmetric. Developing Cross-modal Information Bounds could help quantify how much information from one modality contributes to another, enabling robust multimodal integration, alignment, and transfer in open settings.
\subsubsection{Self-adaptive Information-aware Agents}
Perhaps the most ambitious direction is creating agents capable of self-quantifying their information state. Such an agent would be aware of its informational limits—how much it knows, how much it ignores, and when it should seek new knowledge. This leads to the concept of a Self-Information-Aware Agent, which can measure and regulate its own informational entropy, dynamically adjust its learning capacity, and maintain safety margins based on information divergence. This self-referential information control may form the foundation for autonomous, self-regulating open-world intelligence.

\subsection{Theoretical Outlook}

These research directions collectively point toward a unifying theory of Provable Information Dynamics—a mathematical and conceptual framework capable of describing how information evolves, interacts, and stabilizes in open systems.
By bridging the gap between uncertainty, adaptation, and causality, information theory could provide the first truly general foundation for intelligent behavior under non-stationary conditions.

Future progress will depend on three key developments:

1. Formalization: Establishing precise definitions of open information quantities (entropy, divergence, risk).

2.Estimation: Designing scalable estimators for temporal and multimodal mutual information.

3.Integration: Embedding these measures into learning systems that can reason, explain, and adapt beyond static environments.

Together, these advances could transform open-world learning from an empirical engineering paradigm into a scientifically grounded theory of information-driven intelligence.

\section{Conclusion}
Open-world Machine Learning (OWML) represents a new frontier in artificial intelligence—one where systems must not only learn from limited, incomplete, or shifting data but must also recognize when the world itself changes.
Traditional learning paradigms—rooted in the closed-world assumption—struggle to handle novelty, uncertainty, and non-stationarity. Information theory provides the missing bridge: a unified theoretical language for quantifying, explaining, and ultimately mastering learning under openness.

Throughout this review, we have shown that information theory connects the fundamental components of open-world intelligence: risk, generalization, rejection, learning, and cognition. It redefines risk as information exposure, generalization as information preservation, rejection as uncertainty maximization, and continual learning as information flow regulation.
Within this view, cognition itself can be understood as an emergent property of adaptive information dynamics—an ongoing process of compressing, retaining, and transforming information in response to environmental change.

By embedding open-world learning within an information-theoretic framework, we move from heuristic adaptation to mathematically grounded reasoning. This shift enables the development of provable learning bounds, dynamic information control, and safe adaptive systems capable of operating in non-stationary, uncertain environments.
Information theory thus acts not only as a descriptive tool but as a generative principle—a foundation for designing learning systems that are both accountable and adaptive.

Looking forward, the evolution of open-world intelligence will depend on advancing this information-driven paradigm.
Future learning systems must understand their own informational limits, reason about unknowns, and dynamically regulate their internal knowledge flow. They must move beyond static optimization and embrace continuous adaptation—balancing what is known, what is learnable, and what remains unknowable. In this way, information theory will guide the transformation from learning systems that merely react to their data toward intelligent agents that reason about information itself.

In summary, information theory offers not just mathematical precision but philosophical coherence:
it unifies the diverse mechanisms of open-world learning into a single, principled narrative—a narrative in which intelligence is nothing more, and nothing less, than the art of managing information in an ever-changing world.

\bibliographystyle{apalike}
\bibliography{main}  






\end{document}